\documentclass[letterpaper, 10 pt, conference]{ieeeconf}  % Comment this line out if you need a4paper

\IEEEoverridecommandlockouts                              % This command is only needed if 
\usepackage{graphics} % for pdf, bitmapped graphics files
\usepackage{epsfig} % for postscript graphics files
\usepackage{amsmath} % assumes amsmath package installed
\usepackage{amssymb}  % assumes amsmath package installed
\usepackage[ruled,linesnumbered]{algorithm2e}
\usepackage{todonotes}
\usepackage{cite}
\usepackage[caption=false,font=footnotesize]{subfig}
\usepackage{multirow}
\usepackage{tabularx}
\usepackage{tablefootnote}
\usepackage{footnote}
\usepackage{dblfloatfix}
\usepackage{float}
\floatstyle{plaintop}
\restylefloat{table}

\usepackage{draftwatermark}% to add watermark

\usepackage{tikz}
\usepackage{textcomp}
\usepackage{lipsum}

\newcommand\copyrighttext{%
  \footnotesize \textcopyright 2012 IEEE. Personal use of this material is permitted.
  Permission from IEEE must be obtained for all other uses, in any current or future
  media, including reprinting/republishing this material for advertising or promotional
  purposes, creating new collective works, for resale or redistribution to servers or
  lists, or reuse of any copyrighted component of this work in other works.}
 \newcommand\copyrightnotice{%
 \begin{tikzpicture}[remember picture,overlay]
 \node[anchor=south,yshift=10pt] at (current page.south) {\fbox{\parbox{\dimexpr\textwidth-\fboxsep-\fboxrule\relax}{\copyrighttext}}};
 \end{tikzpicture}%
 }

\title{\LARGE \bf
Sclera Force Control in Robot-assisted Eye Surgery: \\Adaptive Force Control vs. Auditory Feedback
}

\author{Ali Ebrahimi, Changyan He, Niravkumar Patel, {\it Member, IEEE}, Marin Kobilarov, Peter Gehlbach, \\{\it Member, IEEE} and Iulian Iordachita, {\it Senior Member, IEEE}
\thanks{*This work was supported by U.S. National Institute of Health under grant number of 1R01EB023943-01 and Research to Prevent Blindness, New York, New York, USA, and gifts by the J. Willard and Alice S. Marriott Foundation, the Gale Trust, Mr. Herb Ehlers, Mr. Bill Wilbur, Mr. and Mrs. Rajandre Shaw, Ms. Helen Nassif, Ms. Mary Ellen Keck, and Mr. Ronald Stiff.}
\thanks{A. Ebrahimi, C. He,  N. Patel, M. Kobilarov, and I. Iordachita are with the Department of Mechanical Engineering and
Laboratory for Computational Sensing and Robotics at the Johns Hopkins University, Baltimore, MD 21218 USA ( e-mail: aebrahi5, changyanhe, npatel89, marin, iordachita@jhu.edu
}%
\thanks{C. He is also with School of Mechanical Engineering and Automation at Beihang University, Beijing, 100191 China.
}%
\thanks{P. Gehlbach is with the Wilmer Eye Institute, Johns
Hopkins Hospital, Baltimore, MD 21287 USA (e-mail: pgelbach@jhmi.edu).
}%
}

\begin{document}

\SetWatermarkText{Accepted for ISMR 2019}
\SetWatermarkScale{0.4}

\maketitle
\copyrightnotice
\thispagestyle{empty}
\pagestyle{empty}

%%%%%%%%%%%%%%%%%%%%%%%%%%%%%%%%%%%%%%%%%%%%%%%%%%%%%%%%%%%%%%%%%%%%%%%%%%%%%%%%
\begin{abstract}

Surgeon hand tremor limits human capability during microsurgical procedures such as those that treat the eye. In contrast, elimination of hand tremor through the introduction of microsurgical robots diminishes the surgeon’s tactile perception of useful and familiar tool-to-sclera forces. While the large mass and inertia of eye surgical robot prevents surgeon microtremor, loss of perception of small scleral forces may put the sclera at risk of injury. In this paper, we have applied and compared two different methods to assure the safety of sclera tissue during robot-assisted eye surgery. In the active control method, an adaptive force control strategy is implemented on the Steady-Hand Eye Robot in order to control the magnitude of scleral forces when they exceed safe boundaries. This autonomous force compensation is then compared to a passive force control method in which the surgeon performs manual adjustments in response to the provided audio feedback proportional to the magnitude of sclera force. A pilot study with three users indicate that the active control method is potentially more efficient.

\end{abstract}

%%%%%%%%%%%%%%%%%%%%%%%%%%%%%%%%%%%%%%%%%%%%%%%%%%%%%%%%%%%%%%%%%%%%%%%%%%%%%%%%

\section{INTRODUCTION}

  \begin{figure}[t!]
  \centering
    \includegraphics[width=\columnwidth]{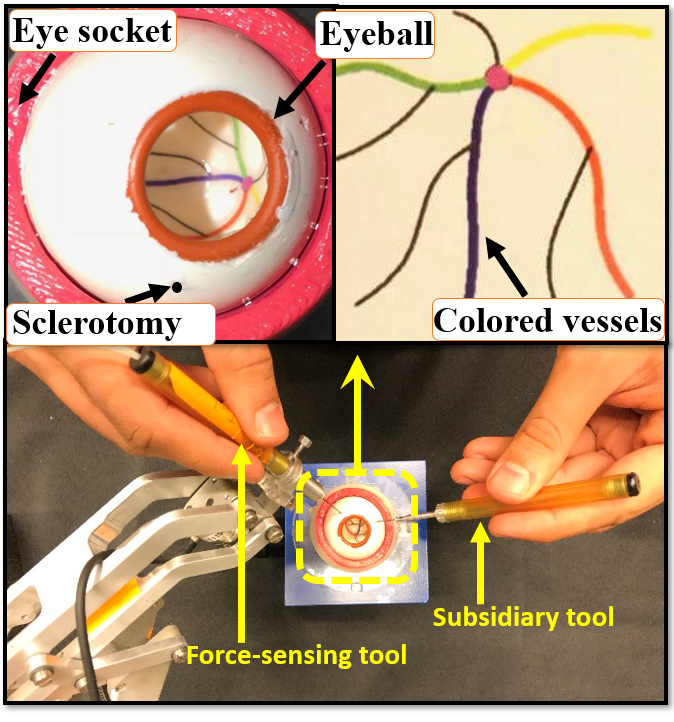}
      \caption{(Bottom) an illustration of how users hold the tools to manipulate the eye. (Top) close-up views of the eyeball and the phantom colored vessels.}
      \label{fg-eyephantom}
\end{figure}

Advancements in robot-assisted eye surgery have successfully reduced hand tremor and have potentially increased patient safety by enhancing tool tip precision during delicate eye surgery maneuvers. Two major categories of robots are presently used for this purpose, collaborative and tele-manipulated. Collaborative robots share control of surgical tool with the surgeon. One such example is the Steady-Hand Eye Robot (SHER) developed at the Johns Hopkins University \cite{uneri2010new}. Using a co-manipulation control strategy and a design based on the Remote-Center-of-Motion (RCM), a 4-DoF (degrees of freedom) robot for retinal surgery was fabricated by Gijbels et al. \cite{gijbels2013design,willekens2017robot}. To obtain an adjustable pivot point, Nasseri et al. designed and developed a compact microsurgical system for ophthalmic surgery which can be mounted on patient's head\cite{nasseri2013introduction,nasseri2013kinematics}. To increase dexterity in intraocular maneuvers, Wei et al. have also developed a hybrid two-armed robotic system  \cite{wei2007design}.

Tele-manipulation robots have emerged as the most clinically ready configurations for retinal surgery at this time. Wilson et al. recently designed and built a novel master-slave intraocular robotic system capable of performing various surgical tasks \cite{wilson2018intraocular}. Also, the University of Tokyo has developed a tele-operated system with which they executed different procedures including vitreous detachment and microcannulation. \cite{ueta2009robot,noda2013impact,tanaka2015quantitative}. Furthermore, a microrobot for performing delicate procedures in retinal surgery was developed by Kummer et al. \cite{kummer2010octomag} which is controlled remotely with a wireless magnetic field.

The first in-human robot-assisted eye surgeries were performed by Edwards et al. \cite{edwards2018first} and Gijbels et al. \cite{gijbels2018human} which can be considered as the most clinically advanced use of robots in eye surgery to date.

In manual eye surgery, surgeons usually rely on the visual and force feedback they attain when they are manipulating the eyeball. However, after introducing robots the interaction forces between the tool shaft and sclera (called sclera force Fig. \ref{fg-tooleye}) are no longer perceived by the surgeon. After doing a user study in robot-assisted eye manipulation, He et al. \cite{he2018user} reported an increase in the average scleral forces if not discerned by the surgeon may put the eyeball at high risk of injury. This would theoretically result from the large inertia and stiffness of the robot relative to the small and delicate tool-to-eye interaction forces. In order to enhance safety, the sensory information related to scleral forces has been restored in prior studies. Ebrahimi et al. \cite{ebrahimi2018real} evaluated the advantage of providing haptic force feedback and audio feedback to surgeons based on sclera force. In another study, Cutler et al. assessed the effect of auditory feedback in restricting tool tip forces from going beyond safe levels in robot-assisted phantom membrane peeling task \cite{cutler2013auditory}.

Based on previous studies, providing audio or haptic feedback to inform surgeons of unsafe levels of force have proven somewhat effective at keeping forces within safe ranges. However, the efficiency of audio or haptic feedback is highly dependent on the surgeon's response to them. As these methods are passive which means robot does not execute any autonomous action to correct unsafe forces. Furthermore, these passive methods of providing feedback may distract a surgeon's attention from the primary tasks of eye surgery because the surgeon should continuously pay attention to them and be prepared to react properly. Therefore, implementing control methods in which the robot performs some or all of the safety actions autonomously may reduce the unsafe forces and prove beneficial in keeping the surgeon focused on primary surgical tasks.

In this paper, we have implemented an adaptive force control method by which the SHER autonomously decreases the magnitude of sclera force when it exceeds safe limits. In addition, we provided auditory feedback based on sclera force to implement a passive sclera force control. Finally, during a user study with three users, the efficiency of the active control of sclera force was compared to the passive audio force feedback.

This paper is organized as following. In section \ref{seactive}, the way the active and the passive sclera force control have been implemented are explained. In section \ref{expsetup}, different components of experimental setup are explained. In section \ref{pilot}, the user study and the way the users are supposed to perform the experiments are delineated. Finally, the comparison results between the active and passive sclera force methods are discussed in section \ref{results}.

%In section III, the adaptive control is used to simulated a simple actuated tool shaft interacting with a 2-D circular eyeball to evaluate the effectiveness of the method in sclera force control. 

%%%%%%%%%%%%%%%%%%%%%%%%%%%%%%%%%%%%%%%%%%%%%%%%%%%%%%%%%%%%%
\section{Active and Passive Sclera Force Control}\label{seactive}

\begin{figure}[t!]
  \centering
    \includegraphics[width=80 mm]{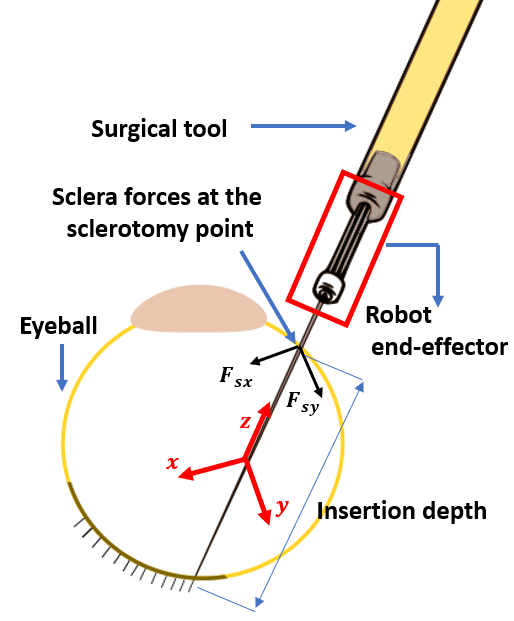}
      \caption{ Schematic diagram for tool and eyball interaction. Sclera force components ($F_{sx}$ and $F_{sy}$) are perpendicular to the tool shaft and expressed in the end-effector coordinate frame (the red coordinate frame whose origin is at a fixed point on the tool shaft and is rigid to the end-effector and the tool). Tool insertion depth is the length of the tool inserted inside the eye.}
      \label{fg-tooleye}
\end{figure}

In this section we explain two different methods (active and passive) deployed to prevent the magnitude of sclera force from exceeding predetermined safe levels. The sclerotomy point (Fig. \ref{fg-eyephantom}) is where the tool shaft leans on the eyeball. It is assumed there is no moment being applied form the sclera to the tool shaft at this point. In addition, the friction force between the sclerotomy and the tool shaft will produce a force component in the $z$ direction of the end-effector (handle) frame (Fig. \ref{fg-tooleye}). This force is assumed to be negligible as well. Hence, only the terms $F_{sx}$ and $F_{sy}$ (depicted in Fig. \ref{fg-tooleye}) will contribute to the magnitude of sclera force, $F_s$ (Fig. \ref{fg-tooleye}). Therefore, the equation for the magnitude of sclera force is written in~\eqref{eq-fs}:
\begin{equation} \label{eq-fs}
\begin{split}
F_s = \sqrt{F_{sx}^2+F_{sy}^2}
\end{split}
\end{equation}

In the active sclera force control strategy, the SHER is made to perform small movements autonomously to diminish $F_s$ when it is going to overstep safe boundaries. This active control strategy is based on an adaptive force control method developed by Roy et. al \cite{roy2002adaptive} for a 1-DoF robot interacting with a compliant environment. We have built upon that control strategy and exploited it in order to make it applicable for 3-D sclera force control in robot-assisted eye surgery.

The second method is a passive control strategy in which we have utilized auditory feedback to warn the users about when $F_s$ reaches unsafe levels. The efficiency of this sound-based sclera force control was already evaluated in \cite{ebrahimi2018real} by our research group. After hearing the warning alarm, the surgeon needs to bring down the sclera force by reorienting the tool direction and executing corrective movements. These corrective movements should be performed simultaneously with the surgical tasks being done. This method is a passive control method because the robot does not perform any autonomous movements to counterbalance the sclera force exceeding safety boundaries. Thus, the surgeon alone adjusts sclera force in response to auditory feedback.

\subsection{Adaptive force control}

\begin{figure}[b!]
  \centering
    \includegraphics[width=70 mm]{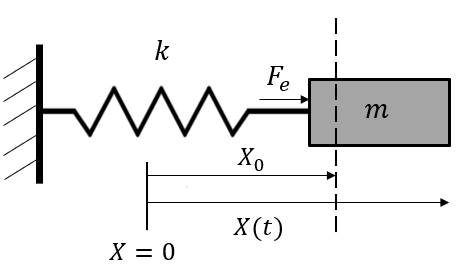}
      \caption{ A 1-DoF robot with mass $m$ interacting with an environment with stiffness $k$}
      \label{fg-1dof}
\end{figure}

In order to control the sclera force with autonomous robot movement, an adaptive control strategy which was originally developed by\cite{roy2002adaptive}, was extended to be applied to the SHER. Considering Fig. \ref{fg-1dof}, the original adaptive control has the following purpose and assumptions:
%\vspace{28pt}

\textbf{Purpose: }
\begin{itemize}
    \item The purpose of the adaptive force control is to make the interaction force ($F_e$) applied to the robot (here the mass $m$ shown in Fig. \ref{fg-1dof}) by the environment with stiffness $k$ to follow a desired pre-defined force $F_d$.
\end{itemize}

\textbf{Assumptions:}
\begin{itemize}
    \item The 1-DoF robot has an inner loop velocity control which makes the robot velocity follow a desired velocity signal $\dot{X}_d$. In other words, the robot is a velocity-controlled robot.
    \item The robot is interacting with an environment with linear compliance $\lambda = \frac{1}{k}$ (based on Fig. \ref{fg-1dof}, $k$ is the environment stiffness) whose quantity is unknown to the force controller. In other words, the relation between the robot displacement $X-X_0$ and the force exerted by the compliant environment, $F_e$, is linear.
\end{itemize}

Roy et al. at \cite{roy2002adaptive} suggests that the control purpose mentioned above will be met under the following control law given in~\eqref{eqadaptivef}.
\begin{equation} \label{eqadaptivef}
\dot{X}_d = \hat{\lambda}\dot{F}_d(t) - \alpha\Delta F(t)
\end{equation}
In the equation above an estimation of the environment compliance $\lambda$ is used because as mentioned the value of $\lambda$ is unknown. This estimation is updated using the following adaptation law:
\begin{equation} \label{eqadaptivef2}
\dot{\hat{\lambda}} = -\Lambda \dot{F}_d(t)\Delta F(t)
\end{equation}

\begin{figure}[t!]
  \centering
    \includegraphics[width=\columnwidth]{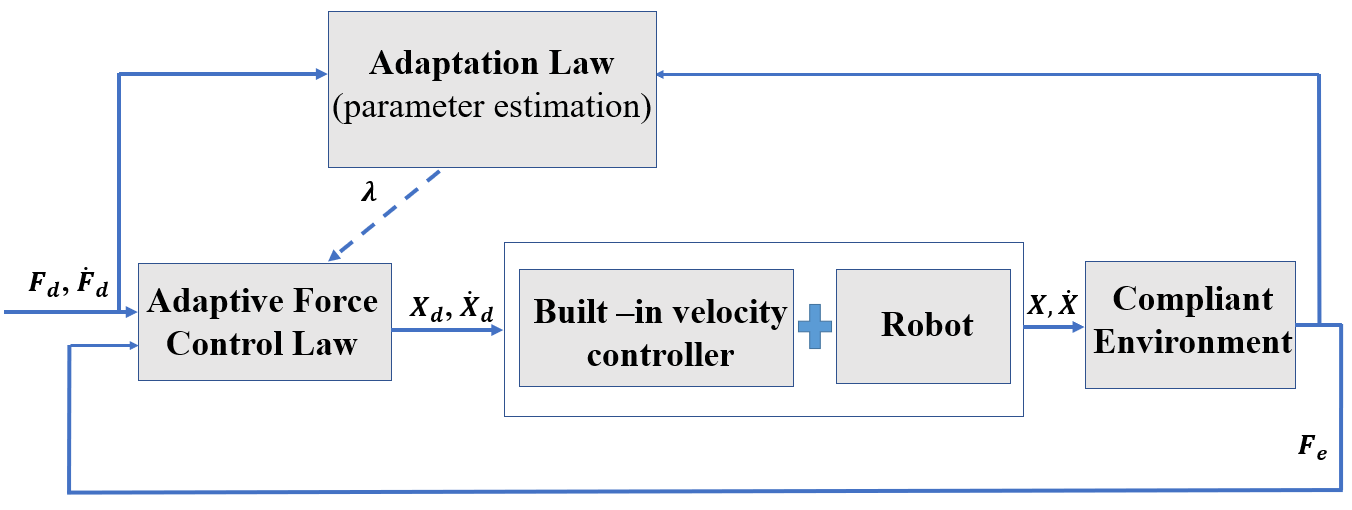}
      \caption{ Block diagram for the 1-DoF adaptive force controller}
      \label{fg-blockdiagram}
\end{figure}

In the equations~\eqref{eqadaptivef} and~\eqref{eqadaptivef2} $\Delta F$ is the error of force control which is the difference between actual force $F_e$ and the desired force $F_d$. The variables $\alpha$ and $\Lambda$ are constant scalars, and $\dot{F}_d(t)$ is the derivative of the desired force signal. As shown in Fig. \ref{fg-blockdiagram} which illustrates the block diagram for the 1-DoF adaptive force control, the signal $\dot{X}_d(t)$ which is produced in~\eqref{eqadaptivef} is fed to the built-in velocity controller of the robot which is assumed to exist. At \cite{roy2002adaptive} it is proved using a Lyapunov function that the control input given in~\eqref{eqadaptivef} and the adaptation law~\eqref{eqadaptivef2} will make the actual force between the robot and the environment, $F_e$, to converge to the desired force $F_d$. We have utilized this control algorithm in robot-assisted eye surgery to control the magnitude of sclera force, $F_s$, when it exceeds safety levels. The idea is to use the adaptive force control above to control each component of sclera force separately. Then, we would be able to make the robot autonomously reduce the magnitude of each component of sclera force ($F_{sx}$ and $F_{sy}$) on desired paths resulting in decreasing $F_s$. 

\begin{figure}[b!]
  \centering
    \includegraphics[width=\columnwidth]{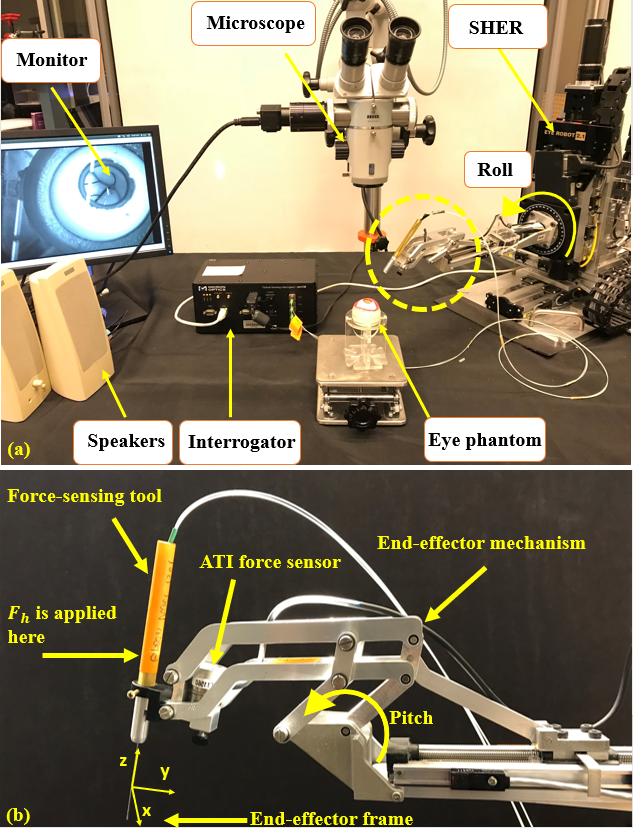}
      \caption{ (a) Experimental setup showing: SHER, FBG interrogator for the force sensing tool, microscope for looking into the eye phantom and speakers to provide the audio feedback. (b) The close-up view for the dashed circle in (a).}
      \label{fg-setup}
\end{figure}

The SHER is a velocity-controlled robot interacting with a compliant environment that fulfills the above assumptions. The environment compliance is unknown to us which justifies the utilization of the adaptive force control. The robot is a 5-DoF manipulator three of which are translational ones which are just motions of the robot base along the Cartesian axis fixed in space. The other two rotational degrees of freedoms are the pitch and roll rotations of the end-effector mechanism which is shown at Fig. \ref{fg-setup}. The end-effector coordinate frame (handle frame) is attached to a fixed point on the tool shaft and is rigid to the end-effector mechanism which is shown in Fig. \ref{fg-setup}-b. 

The normal impedance control of the robot makes it move in accordance with the forces and torques applied by the surgeon to the tool handle (the point where the surgeon grabs the tool, Fig. \ref{fg-setup}-b. This vector of forces and torques is denoted by $F_h^b\in \mathbb{R}^6 $ which the superscript $b$ indicates that the vector is projected in the handle frame explained above. Thus, for the normal impedance control of the robot, the desired translational and rotational velocities ($\dot{X}_d^b$) of the end-effector frame is produced based on~\eqref{eqimpedance}. The SHER is a velocity-conterolled robot and the embedded velocity controller of the robot is able to make the robot end-effector frame follow any bounded desired velocity vector $\dot{X}_d^b$.
\begin{equation} \label{eqimpedance}
\dot{X}_d^b = \mathbb{K} F_h^b 
\end{equation}
In~\eqref{eqimpedance}, the matrix $\mathbb{K}\in \mathbb{R}^{6\times6}$ is a constant diagonal gain matrix. The first three elements of $\dot{X}_d^b$ are the linear velocity of handle frame origin and the last three elements of $\dot{X}_d^b$ are the angular velocities of the handle frame. Hence, each of the six elements of $\dot{X}_d^b$ is related to its corresponding value in the force vector $F_h^b$ which means the robot obeys the way the surgeon wishes to manipulate the SHER. 

In order to augment the sclera safety during eye manipulation, the adaptive controller~\eqref{eqadaptivef} is integrated into~\eqref{eqimpedance} to generate the first two components of $\dot{X}_d^b$ which pertain to the $x$ and $y$ linear velocities of the handle frame origin. Then we can define decreasing desired reference signals for $F_{sx}$ and $F_{sy}$ to be followed. In other words, whenever $F_s$ exceeds safe levels, the magnitude of both components of sclera force will be reduced based on the desired paths designed for the $F_{sx}$ and $F_{sy}$ using the adaptive force control method. Decreasing the magnitude of both components of sclera force will result in decreasing the sclera force magnitude $F_s$ which has been the goal of active control of sclera force norm.

To encapsulate, equation set~\eqref{eqadaptivesclera} are used whenever the autonomous norm control of sclera force is switched on (instead of using~\eqref{eqimpedance}). The scalar $\dot{X}_d^b[i]$ is the $i\textsuperscript{th}$ entry of the vector $\dot{X}_d^b$.
\begin{equation} \label{eqadaptivesclera}
\begin{split}
&\dot{X}_d^b[1] = \hat{\lambda}_1\dot{F}_{dx}(t) - \alpha_1\Delta F_x(t) \\
&\dot{X}_d^b[2] = \hat{\lambda}_2\dot{F}_{dy}(t) - \alpha_2\Delta F_y(t) \\
&\dot{X}_d^b[j] = \mathbb{K}[jj] F_h^b[j] \quad \text{for} \quad j=3,4,5,6
\end{split}
\end{equation}
Where ${F}_{dx}$ and ${F}_{dy}$ are the desired reference signals for ${F}_{sx}$ and ${F}_{sy}$, respectively. The terms $\Delta F_x$ and $\Delta F_y$ are also the tracking error for sclera force components which are ${F}_{sx}-{F}_{dx}$ and ${F}_{sy}-{F}_{dy}$, respectively. For the first two equations in~\eqref{eqadaptivesclera}, the adaptive laws to update the estimation for the environment compliance along the $x$ and $y$ directions of the body frame are given in ~\eqref{eqadaptivesclera2}.
\begin{equation} \label{eqadaptivesclera2}
\begin{split}
&\dot{\hat{\lambda}}_1 = -\Lambda_1 \dot{F}_{dx}(t)\Delta F_x(t) \\
&\dot{\hat{\lambda}}_2 = -\Lambda_2 \dot{F}_{dy}(t)\Delta F_y(t) 
\end{split}
\end{equation}
Thus, when $F_s$ goes beyond the upper safe level $L$, robot control equations are switched from~\eqref{eqimpedance} to~\eqref{eqadaptivesclera} for autonomous reduction of sclera force. This, of course, does not stops the surgeon's manipulation because the last four elements of $\dot{X}_d^b$ are still produced based on the surgeon's interaction force $F_h$ as it is apparent from~\eqref{eqadaptivesclera}.

The decreasing desired reference signals for $F_{sx}$ and $F_{sy}$ are written in~\eqref{eqreferences}.
\begin{equation} \label{eqreferences}
\begin{split}
&F_{dx} = \frac{F_{sx}^0}{2}(e^{-a(t-t_0)}+1) \\
&F_{dy} = \frac{F_{sy}^0}{2}(e^{-a(t-t_0)}+1) 
\end{split}
\end{equation}
where $t_0$ is the time when $F_s$ reaches the upper safe bound of $L$ and $a$ is also a positive constant scalar representing how fast the desired trajectory falls. The scalars $F_{sx}^0$ and $F_{sy}^0$ are the values of $F_{sx}$ and $F_{sy}$ at time $t=t_0$. By little inspection, it is apparent that both of the desired trajectories defined above will make the magnitudes of $F_{sx}$ and $F_{sy}$ to decrease. Because, for example if $F_{sx}^0$ is positive, $F_{dx}$ will be a decreasing signal and if $F_{sx}^0$ is negative $F_{dx}$ will be an increasing signal. Thus, by these desired trajectories for $F_{sx}$ and $F_{sy}$, $F_s$ will also decrease and the safety of sclera force magnitude in robot-assisted eye manipulation will be guaranteed. 
If we keep the adaptive set of control equations~\eqref{eqadaptivesclera}, the magnitude of $F_{sx}$ and $F_{sy}$ will continuously decrease and the $t= \infty$ they will reach the constant values of $\frac{F_{sx}^0}{2}$ and $\frac{F_{sy}^0}{2}$, respectively. However, we want to switch back to the normal impedance control of the robot~\eqref{eqimpedance} as soon as possible since this control completely obeys the force $F_h$ and feels more convenient to the surgeon. Thus, the control equations of~\eqref{eqimpedance} will again be applied when $F_{sx}$ and $F_{sy}$ reach $\frac{3F_{sx}^0}{4}$ and $\frac{3F_{sy}^0}{4}$ respectively which happens at a short finite time.

\subsection{\vspace{-0.02cm}Auditory Substitution}
For the purpose of passive sclera force control, auditory warning alarms in the form of beeps are provided to the surgeon when $F_s$ reaches unsafe levels. As the surgeon hears the alarms, s/he should do corrective movements while s/he is performing surgical tasks to bring down the sclera force and get rid of the warning alarm. This method was used by our group in \cite{ebrahimi2018real}, and the advantage of providing audio feedback in enhancing safety was investigated. Thus, for passive control of scelra force the surgeon does the eye manipulation while the impedance control equation set~\eqref{eqimpedance} is always implemented which means there is no switch to the equation set~\eqref{eqadaptivesclera}. 

The first warning noise is sounded when $F_s>L_1$, and it is a low volume and low frequency noise. The second level of warning alarm is produced when $F_s>L_2$. When $F_s$ reaches $L_3$ which is determined to be the unsafe level for $F_s$, a constant noise with relatively high volume is emitted to inform the surgeon of the safe bounds being passed. Thus, the first two emitted sounds produced at $L_1$ and $L_2$ act as a precaution for the unsafe level $L_3$ which is potentially harmful to sclera tissue. 

\section{Experimental Setup and Method}\label{expsetup}
To conduct experiments a setup which is depicted in Fig. \ref{fg-setup} is prepared. The user should look though the microscope and hold the surgical tool which is also grasped by the robot to manipulate the eyeball. The eye phantom is placed in a 3-D printed eye socket. To measure the real-time value of $F_{sx}$ and $F_{sy}$, a dual force-sensing tool which is built and calibrated based on \cite{he2014multi} is used (see section \ref{DualTool}). Furthermore, the ATI force sensor (Fig. \ref{fg-setup}-b) is attached to the robot handle which measures the vector $F_h^b$. To produce sound for the passive control method, two speakers are placed close to the user. Using the TCP/IP connection, data is transmitted between different modules of the system. Sclera force measurements, robot position, velocity and other required data are collected and manipulated through the software package (developed using the CISST framework, a collection of libraries for development of computer-assisted intervention systems; InfinityQS, Fairfax). 
The unsafe upper bound level of sclera force was set to be $120$ mN according to what is explained in \cite{ebrahimi2018real}. Therefore, we put the levels of $L_1$, $L_2$ and $L_3$ which were defined in the passive control of sclera force to be $80$ mN, $100$ mN and $120$ mN, respectively. Thus, a soft alarm will be sounded at $80$ mN, and it gradually merges into higher and rough warning noise when $F_s$ reaches $120$ mN.

For passive sclera control the warning alarms starts at $80$ mN and progressively increases as we reach $120$ mN. To have a fair comparison between the active and passive sclera control methods the same precaution idea is also applied for the active sclera control before reaching the unsafe upper bound of $120$ mN. Thus, the adaptive control method is switched on when $F_s$ reaches $L=100$ mN. The control parameters are also specified as following. In ~\eqref{eqimpedance} the diagonal matrix $\mathbb{K}$ is set to be $7.5\mathbb{I}$ where $\mathbb{I}$ is the identity matrix. In~\eqref{eqadaptivesclera} and~\eqref{eqadaptivesclera2}, the scalars $\alpha_1$, $\alpha_2$, $\Lambda_1$ are $\Lambda_2$ set to be $0.2$, $0.2$, $5\times10^{-6}$ and $5\times10^{-6}$, respectively. To decide on the value of the parameter $a$ in~\eqref{eqreferences} we did some experiments to find an optimized value. We saw that for high values ($a>>1$) the robot was acting in a fast and unstable way. In contrast, for low values of $a$ sclera forces were not following the desired trajectories well. Eventually,  $a$  was set to $1$ to have a trade-off between stability and accuracy in force tracking error.

\subsection{Steady-Hand Eye Robot} \label{SteadyHand}
The Steady-Hand Eye Robot is a cooperatively controlled 5-DoF robot for microsurgical manipulation in eye surgery that helps surgeons reduce hand tremor and have more stable, smooth and precise manipulations. As mentioned before, the SHER is a velocity-controlled robot and the velocity of the motors of each joint is controlled by an embedded velocity controller (Galil 4088, Galil, 270 Technology Way, Rocklin, CA 95765).

\subsection{Dual Force-Sensing Tool} \label{DualTool}
To measure sclera forces $F_{sx}$ and $F_{sy}$, FBG optical strain sensors were used as explained by \cite{he2014multi}. The FBG sensors are very sensitive to strain which makes them suitable for applications measuring mN-order sclera forces. Three optical fibers were attached around a 25-gauge Nitinol wire at a separation angle of $120 ^\circ$ from each other (see \cite{he2014multi}). The FBG fibers are connected to an optical sensing interrogator (sm130-700 from Micron Optics Inc., Atlanta, GA) which sends the FBG raw optical data with a maximum frequency of 1 KH to the computer to calculate real-time force data (Fig. \ref{fg-setup}). In addition to measuring the components of sclera force, this tool design enables measuring tool tip force and the scalar value of insertion depth, as well. In this paper we utilize only the sclera force and we do not utilize the two other measured variables.

The calibration and validation processes for the tool measurement precision were carried out as previously reported by \cite{he2014multi}. The mean square error for the sclera force and insertion depth measurements were calculated to be $3.8$ mN and $0.29$ mm, respectively.

\subsection{Eye Phantom} \label{EyePhantom}

The artificial silicon-made eye phantom is placed into a 3D-printed socket as shown in Fig. \ref{fg-eyephantom}. To produce a realistic friction coefficient between the eye phantom and the socket, the surface between them is lubricated with mineral oil. In addition, phantom vessels which are just colored curves are printed on the posterior portion of the eyeball's interior.

\section{Pilot Study} \label{pilot}
In order to compare the efficacy of active and passive sclera force control methods in keeping the sclera force magnitude $F_s$ below the unsafe level of $120$ mN, we conducted a pilot study with three right-handed users. The research study attained the approval from the Johns Hopkins University Institutional Review Board for conducting user studies. 
The users were asked to hold the force-sensing tool attached to the robot with their right hand and move it collaboratively with the SHER to follow four colored vessels (Fig. \ref{fg-eyephantom}) with the tool tip. A secondary but not-force-sensing tool is held in the left hand to facilitate the task (Fig. \ref{fg-eyephantom}). However, we asked the users to use their right hand as the primary manipulation element. In \cite{he2018bimanual}, it is shown that in bimanual eye manipulation the dominant hand (here the right hand) applies significantly higher sclera forces.

Each user performed $10$ trials for active sclera force control and $10$ trials for passive sclera force control. In each trial, the permutation of the colored vessels to be followed is changed in a random way. During the manipulationو the users were asked to view the eye through the microscope as a surgeon would in surgery. It is notable to say that user 1 was the most experienced while user 3 was the least experienced in operating the SHER.

\section{Results and Discussion}\label{results}

In order to compare the two methods of sclera force control, some characteristic variables are defined. The first, is the total time required to complete the task of following four vessels. In order to quantify the augmentation of sclera force safety, the time spent with forces exceeding more than the upper safe boundary, $120$ mN, is utilized. Also, the average value of sclera force and the maximum probable sclera force which is the force near where most of the time spent occurred during vessel following are calculated. 

\begin{figure}[t!]
  \centering
    \includegraphics[width=\columnwidth]{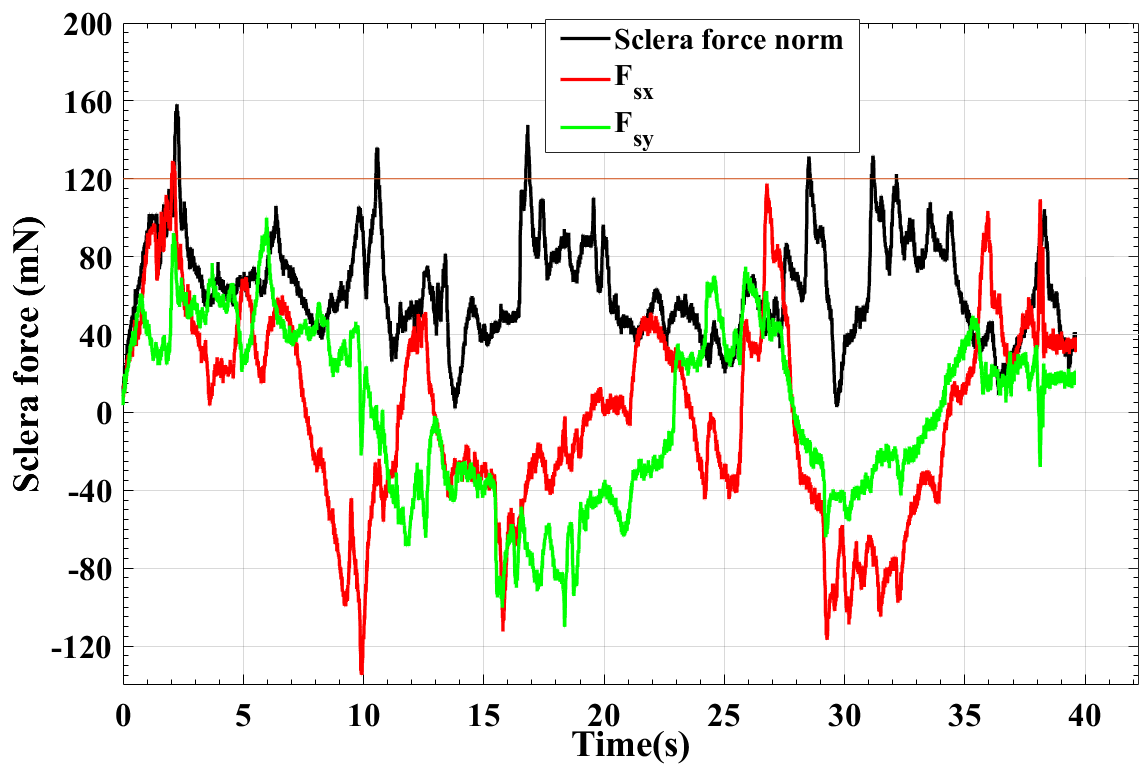}
      \caption{A characteristic plot for $F_{sx}$, $F_{sy}$ and $F_s$ of one of the 10 trials done by user 3. For this trial the adaptive norm control of sclera force was implemented.}
      \label{fg-adaptivenorm}
\end{figure}

\begin{figure}[t!]
  \centering
    \includegraphics[width=\columnwidth]{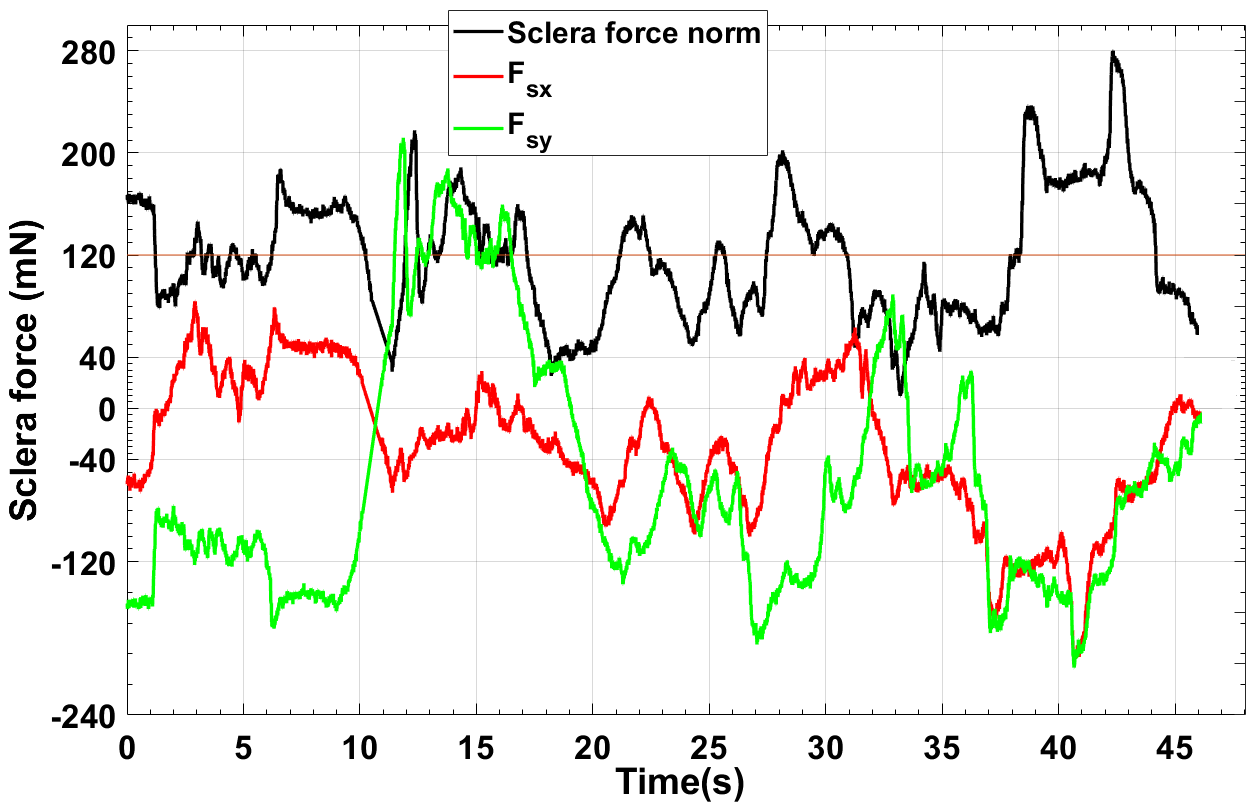}
      \caption{A characteristic plot for $F_{sx}$, $F_{sy}$ and $F_s$ of one of the 10 trials done by user 3. For this trial the passive control of sclera force was implemented.}
      \label{fg-audionorm}
\end{figure}

A typical trial for active control and audio feedback control of sclera force are depicted in Figs. \ref{fg-adaptivenorm} and \ref{fg-audionorm}, respectively. These figures just represents one of the trials done by user 3 which as mentioned before was the most novice user in working with the SHER. 

% \begin{figure}[t!]
%  \centering
 %   \includegraphics[width=\columnwidth]{table.png}
%      \caption{Average results for each of the three users in the pilot study. For each user the %results are averaged over the 10 trials done by the user}
%      \label{fg-table}
%\end{figure}
\begin{table}[t]
\centering
    \caption{ \footnotesize Average results for each of the three users in the pilot study. For each user the results are averaged over the 10 trials done by the user for that specific task. The numbers in parenthesis are standard deviation}
    \includegraphics[width=\columnwidth]{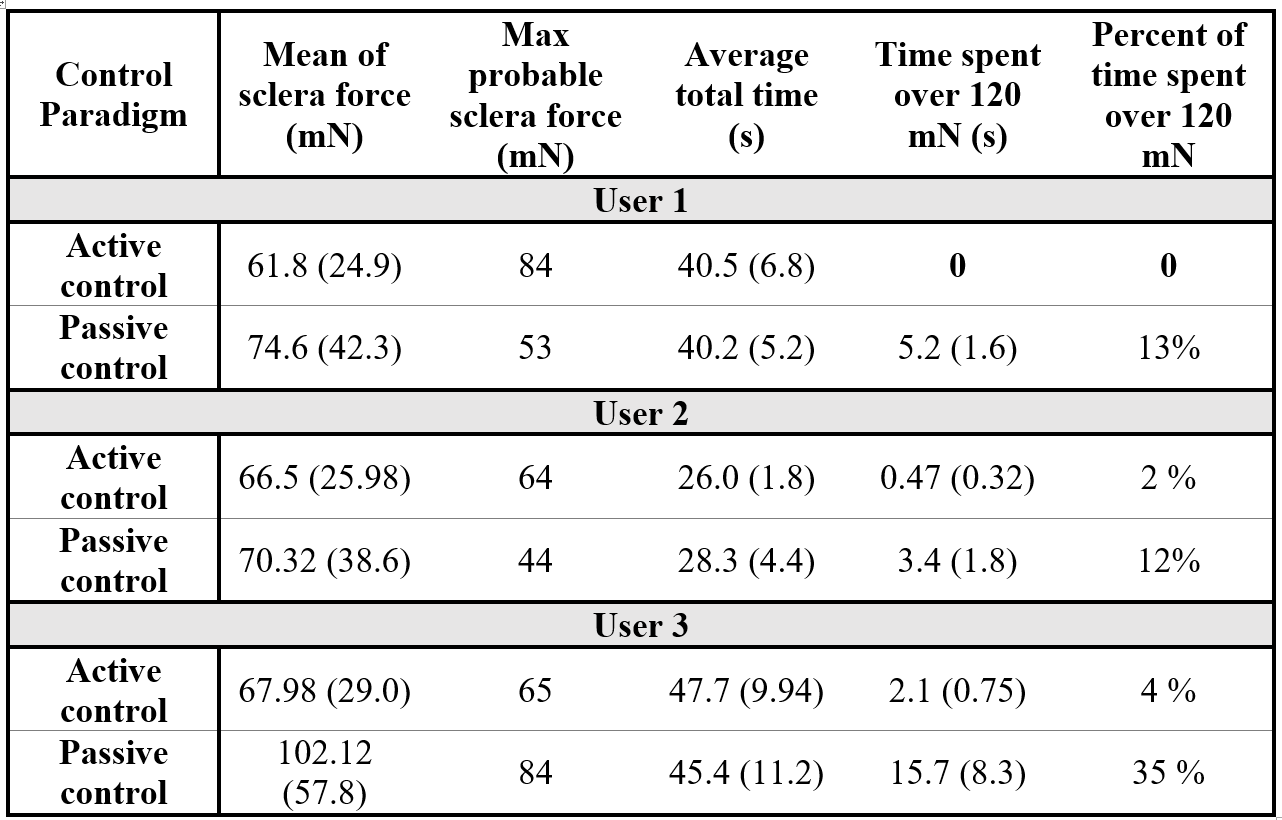}
      \label{table}
\end{table}
As it is shown in Fig.  \ref{fg-adaptivenorm}, the robot is controlling the components of sclera force as the $F_s$ exceeds $100$ mN by decreasing the norm of $F_{sx}$ and $F_{sy}$. It can be seen that $F_s$ is mostly below the unsafe level, $120$ mN, and in some cases it has a spike and goes in excess of the unsafe boundary, but the robot immediately brings it back to the safe region. However, if we look into the passive control method results in Fig. \ref{fg-audionorm}, it takes some time for the user to react to the warning alarm stating the unsafe level has been exceeded. Thus, more time is spent on forces more than $120$ mN. Furthermore, another shortcoming of the audio feedback method is that, the user should always pay attention if an alarm is sounded and react to that promptly. This, may distract the user or in reality the surgeon's attention from the primary and absolutely delicate tasks of eye surgery. However, the active control method not only more enhances the safety of sclera force, but also does not disrupt the surgical tasks done by the surgeon and thus is always taking care of the sclera force norm during the surgery.

By looking into the results of Table \ref{table}, it is observed that the time spent over $120$ mN in active control method is significantly less than that for the audio norm control for all users. In other words, the active control method is acting more efficiently in preventing the sclera force from going beyond unsafe levels. User 1 who has the most experience in working with the SHER has zero time spent over $120$ mN. For the most novice user which is user number 3, the ratio of the averaged time spent over $120$ mN to the averaged total time of the experiment is as low as $4\%$ for the active control method which means the sclera force in $96\%$ of time is located in safe regions. Also, for all of the users, the average value of sclera force in the active control method is less than that quantity for the passive sclera force control. 

Thus, active sclera force control enhances sclera force safety more than the other method without having the disadvantage of distracting the surgeon's attention. One drawback of the active sclera control is that the first two elements $\dot{X}_d^b$ are not obeying the user's interaction force $F_h$ (according to~\eqref{eqadaptivesclera}) which means the user does not have as much control over the robot's movements as compared to passive control method which has~\eqref{eqimpedance} as the governing control equations. However, the other last four elements of $\dot{X}_d^b$ are still abiding by $F_h$ according to~\eqref{eqadaptivesclera} which means the robot is mostly obeying the user commands rather than paying attention to the adaptive sclera force control. Thus, the user does not appreciate that the robot inhibits his/her commands. Nevertheless, user 3 stated that when the adaptive control was switched on, the robot deviated a little from the direction he intended to move along. Thus, the control parameters require further optimization in order to impart minimal impedance to the surgeon while protecting the eye from scleral force injury. Finally, we did not observe any instability during the experiments due to switching between control methods, but one of our future goals is to further investigate the stability of the switching system.

\vspace{12pt}
%\color{red}
\bibliographystyle{IEEEtran}
\bibliography{eyebrp}
\end{document}